\documentclass[11pt,a4paper]{article}
\usepackage[hyperref]{acl2019}
\usepackage{times}
\usepackage{latexsym}
\usepackage{booktabs}
\usepackage{multirow}
\usepackage{url}
\usepackage{graphicx}
\usepackage{amsmath,amsfonts,amssymb}

\usepackage{xcolor}

\aclfinalcopy \def\aclpaperid{157}

\title{Low-Level Linguistic Controls for Style Transfer and Content Preservation}

\newcommand{\csss}{\large $\boldsymbol{\flat}$}
\newcommand{\litss}{\large $\boldsymbol{\sharp}$}

\author{Katy Ilonka Gero\textsuperscript{\csss}\thanks{\enspace Equal contribution.}
~~ Chris Kedzie\textsuperscript{\csss}\footnotemark[1]
~~ Jonathan Reeve\textsuperscript{\litss}~~ Lydia B. Chilton\textsuperscript{\csss}\\
Columbia University\\
\textsuperscript{\csss}Dept. of Computer Science, \textsuperscript{\litss}Dept. of English and Comparative Literature \\
\texttt{katy@cs.columbia.edu}, ~ \texttt{kedzie@cs.columbia.edu}, \\
\texttt{jpr2152@columbia.edu}, ~ \texttt{chilton@cs.columbia.edu}
}

\date{}
\begin{document}
\maketitle
\begin{abstract}
Despite the success of style transfer in image processing, it has seen
limited progress in natural language generation. Part of the problem is that 
content is not as easily decoupled from style in the text domain. Curiously, in the  field of stylometry, content does \emph{not} figure
prominently in practical methods of discriminating stylistic elements, such as authorship and genre.
Rather, syntax and function words are the most salient features.
Drawing on this work, 
we model style as a suite of low-level linguistic controls, such as 
frequency of pronouns, prepositions, and subordinate clause constructions. 
We train a neural encoder-decoder model to reconstruct reference sentences given
only content words and the setting of 
the controls. 
We perform style transfer by keeping the content words fixed while
adjusting the controls to be indicative of another style. 
In experiments, we show that the model reliably responds to the linguistic controls and perform both automatic and manual evaluations on style transfer.
We find we can fool a style classifier 84\% of the time, and 
that our model produces highly diverse and stylistically distinctive outputs.
This work introduces a formal, extendable model of style that can add control to any neural text generation system.
    
\end{abstract}

\section{Introduction}

All text has style, whether it be formal or informal, polite or aggressive, colloquial, persuasive, or even robotic. Despite the success of style transfer in image processing \cite{gatys2015neural, imagetransfer}, there has been limited progress in the text domain, where
disentangling style from content is particularly difficult.

To date, most work in style transfer relies on the availability of meta-data, such as sentiment, authorship, or formality. While  meta-data can provide insight into the style of a text, it often conflates style with content, limiting the ability to perform style transfer while preserving content.
Generalizing style transfer requires separating style from the meaning of the text itself. 
The study of literary style can guide us. 
For example, in the digital humanities and its subfield of stylometry, content doesn't figure prominently in practical methods of discriminating authorship and genres, which can be thought of as style at the level of the individual and population, respectively. Rather, syntactic and functional constructions are the most salient features.

In this work, we turn to literary style as a test-bed for style transfer, and build on work from literature scholars using computational techniques for analysis. In particular we draw on stylometry: the use of surface level features, often counts of function words, to discriminate between literary styles.
Stylometry first saw success in attributing authorship to the disputed Federalist Papers \cite{MostellerInferenceDisputedAuthorship2007}, but is recently used by scholars to study things such as the birth of genres \cite{underwood2016life} and the change of author styles over time \cite{reeve2019}.
The use of function words is likely not the way writers intend to express style, but they appear to be downstream realizations of higher-level stylistic decisions.

We hypothesize that surface-level linguistic features, such as counts of personal pronouns, prepositions, and punctuation, are an excellent definition of literary style, as borne out by their use in the digital humanities, and our own style classification experiments.  We propose a controllable neural encoder-decoder model in which these features are modelled explicitly as decoder feature embeddings. In training, the model learns to reconstruct a text using only the content words and the linguistic feature embeddings. We can then transfer arbitrary content 
words to a new style without parallel data by setting the low-level style feature embeddings to be indicative of the target style.

This paper makes the following contributions:

\begin{itemize}
    \item A formal model of style as a suite of controllable, low-level linguistic features
    that are independent of content.
        \item An automatic evaluation showing that our model fools a style classifier 84\% of the time.
        \item A human evaluation with English literature experts, including recommendations for dealing with the entanglement of content with style. \end{itemize}

\section{Related Work}

\subsection{Style Transfer with Parallel Data}

Following in the footsteps of machine translation, style transfer in text has seen success by using parallel data. \citet{shakespeare} use modern translations of Shakespeare plays to build a modern-to-Shakespearan model. \citet{formality-style} compile parallel data for formal and informal sentences, allowing them to successfully use various machine translation techniques. 
While parallel data may work for very specific styles, the difficulty of finding parallel texts dramatically limits this approach.

\subsection{Style Transfer without Parallel Data}

There has been a decent amount of work on this approach in the past few years \cite{zhaoetal, fu2018style}, mostly focusing on variations of an encoder-decoder framework in which style is modeled as a monolithic style embedding. The main obstacle is often to disentangle style and content. However, it remains a challenging problem.

Perhaps the most successful is \citet{multipleattribute}, who use a de-noising auto encoder and back translation to learn style without parallel data. 
\citet{styleopinion} outline the benefits of automatically extracting style, and suggest there is a formal weakness of using linguistic heuristics. In contrast, we believe that monolithic style embeddings don't capture the existing knowledge we have about style, and will struggle to disentangle content.

\subsection{Controlling Linguistic Features}

Several papers have worked on controlling style when generating sentences from restaurant meaning representations \cite{mr:oraby, deriu2018syntactic}. 
In each of these cases, the diversity in outputs is quite small given the constraints of the meaning representation, style is often constrained to interjections (like ``yeah''), and there is no original style from which to transfer.

\citet{ficler} investigate using stylistic parameters 
and content parameters 
to control text generation using a movie review dataset. Their stylistic parameters are created using word-level heuristics and they are successful in controlling these parameters in the outputs. Their success 
bodes well for our related approach in a style transfer setting, in which the content (not merely content parameters) is held fixed.

\subsection{Stylometry and the Digital Humanities}
Style, in literary research, is anything but a stable concept, but it nonetheless has a long tradition of study in the digital humanities. In a remarkably early quantitative study of literature, \citet{mendenhall_characteristic_1887} charts sentence-level stylistic attributes specific to a number of novelists.
Half a century later, \citet{fucks1952} builds on earlier work in information theory by \citet{shannon1948mathematical}, and defines a literary text as consisting of two ``materials": ``the \emph{vocabulary}, and some structural properties, the \emph{style}, of its author." 

Beginning with \citet{MostellerInferenceDisputedAuthorship2007}, statistical approaches to style, or stylometry, join the already-heated debates over the authorship of literary works.
A noteable example of this is the ``Delta" measure,
which uses z-scores of function word frequencies \cite{burrows2002delta}. 
\citet{craig2009shakespeare} find that Shakespeare added some material to a later edition of Thomas Kyd's \emph{The Spanish Tragedy}, and that Christopher Marlowe collaborated with Shakespeare on \emph{Henry VI}. 

\section{Models}
\subsection{Preliminary Classification Experiments}

The stylometric research cited above suggests that the most frequently
used words, e.g. function words, are most discriminating of authorship and literary style.\footnote{Curiously, these are most often the kinds of words that are manually
removed for text classification.} We investigate these claims using three 
corpora that have distinctive styles in the literary community: gothic novels, philosophy books, and pulp science fiction, hereafter sci-fi.  

We retrieve gothic novels and philosophy books from Project Gutenberg\footnote{\url{www.gutenberg.org}} and pulp sci-fi from Internet Archive's Pulp Magazine Archive\footnote{Specifically, Robin Sloan's OCR'ed corpus: \url{https://archive.org/details/scifi-corpus}}. We 
partition this corpus into train, validation, and test sets the sizes of which can be found in Table \ref{table:corpora}.

\begin{table}
\center
\begin{tabular}{lccc}
\toprule
            & Train         & Dev           & Test \\
Style       &  \small{Words/Sent}   & \small{Words/Sent}    & \small{Words/Sent} \\
\midrule
Sci-fi      & 7.1M/344k     & .9M/43k       & .9M/43k \\
Phil        &  1.2M/120k    & .15M/15k      & .15M/15k \\
Gothic      &  .4M/74k      & .05M/9k       &.05M/9k \\
\bottomrule
\end{tabular}
\caption{The size of the data across the three different styles investigated.}
\label{table:corpora}
\end{table}

In order to validate the above claims, we train five different classifiers to predict the literary 
style of sentences from our corpus. Each classifier has gradually more content words replaced
with part-of-speech (POS) tag placeholder tokens.
The \textit{All} model is trained on sentences with all proper nouns replaced by `PROPN'.
The models \textit{Ablated N, Ablated NV}, and  \textit{Ablated NVA} replace nouns, nouns \& verbs,
and nouns, verbs, \& adjectives with the corresponding POS tag respectively.
Finally, \textit{Content-only} is trained on sentences with all words that are not tagged as NOUN, VERB, ADJ removed; the remaining words are not ablated.

We train the classifiers on the training set, balancing the class distribution to make sure there are the same number of sentences from each style. Classifiers are trained using fastText \cite{fasttext}, using tri-gram features with all other settings as default.
\autoref{table:classifiers} shows the accuracies of the classifiers.

\begin{table}
\center
\begin{tabular}{lrrrr}
\toprule
Classifier	        	&all	&scifi	&goth	&phil \\
\midrule
All            	&0.86	&0.86	&0.87	&0.84\\
Content only    &0.80	&0.78	&0.80	&0.84\\
Ablated N		&0.81	&0.80	&0.85	&0.83\\
Ablated NV		&0.80	&0.83	&0.77	&0.72\\
Ablated NVA		&0.75	&0.73	&0.72	&0.80\\
\bottomrule
\end{tabular}
\caption{Accuracy of five classifiers trained using trigrams with fasttext, for all test data and split by genre. Despite  heavy ablation, the \textit{Ablated NVA} classifier has an accuracy of 75\%, suggesting synactic and functional features alone can be fully predictive of style.}
\label{table:classifiers}
\end{table}

The styles are highly distinctive: the \textit{All} classifier has an accuracy of 86\%. Additionally, even the \textit{Ablated NVA} is quite successful, with 75\% accuracy, even without access to any content words. The \textit{Content only} classifier is also quite successful, at 80\% accuracy. This indicates that these stylistic genres are distinctive at both the content level and at the syntactic level.

\subsection{Formal Model of Style}
\label{sec:formalstyle}

Given that non-content words are distinctive enough for a classifier to determine 
style, we propose a suite of low-level linguistic feature counts (henceforth, controls) as our formal, content-blind
definition of style. The style of a sentence is represented as a vector
of counts of closed word classes (like personal pronouns) as well as counts of syntactic features like the number of SBAR 
non-terminals in its constituency parse, since clause structure has
been shown to be indicative of style \cite{scaleofsentence}.
Controls are extracted heuristically, and almost all rely on counts of pre-defined word lists. For constituency parses we use the 
Stanford Parser \cite{stanfordcore}.
\autoref{table:controlexamples} lists all the controls along with examples.

\begin{table}
\center
\begin{tabular}{lll}
\toprule
Control &  Source &  Example \\
\midrule
S             &  parse &      n/a  \\
SBAR          &  parse &      n/a  \\
ADVP          &  parse &     n/a \\
FRAG          &  parse &      n/a \\
conjunction   &  word list &  and, or, yet, but \\
determiner    &  word list &  the, an, this \\
3rdNeutralPer &  word list &  they, their, it \\
3rdFemalePer  &  word list &  she, her \\
3rdMalePer    &  word list &  he, his \\
1stPer        &  word list &  I, my, we \\
2ndPer        &  word list &  you, your \\
3rdPer        &  word list &  they, she, he \\
helperVerbs   &  word list &  be, am, could   \\
negation      &  word list &  no, not \\
simple prep   &  word list &  for, despite     \\
position prep &  word list &  above, down   \\
punctuation   &  word list &  , ; : - \_ ( \\
\bottomrule
\end{tabular}
\caption{All controls, their source, and examples. Punctuation doesn't include end punctuation.}
\label{table:controlexamples}
\end{table}

\subsubsection{Reconstruction Task} Models are trained with a reconstruction task, in which a distorted version of a reference sentence is input and the goal is to output the original reference. 

\begin{figure}[t]
    \centering
    \includegraphics[width=.48\textwidth]{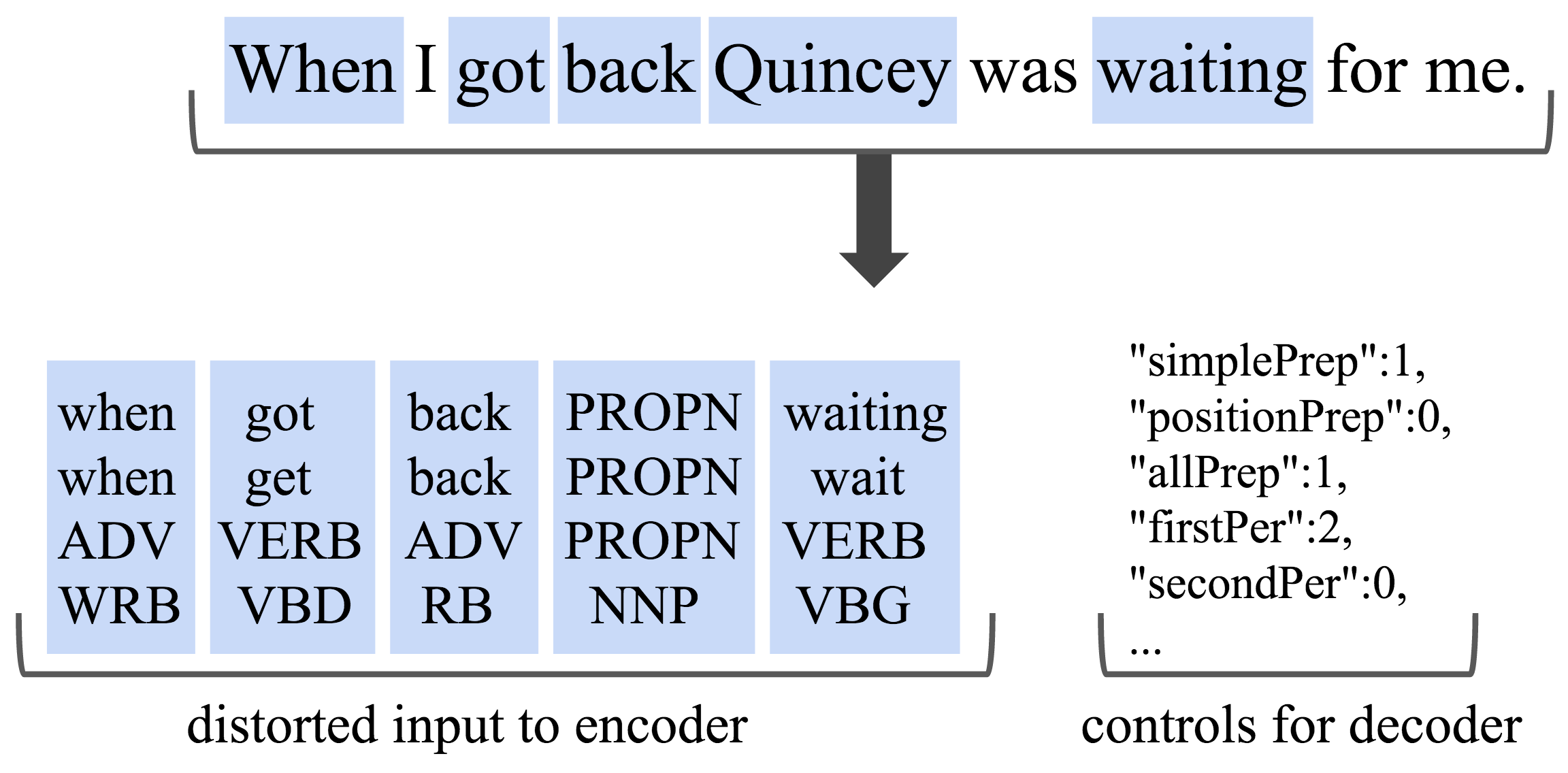}
    \caption{How a reference sentence from the dataset is prepared for input to the model. Controls are calculated heuristically, and then removed from the sentence. The remaining words, as well as their lemmatized versions and part-of-speech tags, are used as input separately.}
    \label{fig:sentenceinput}
\end{figure}

\autoref{fig:sentenceinput} illustrates the process. Controls are calculated heuristically. All words found in the control word lists are then removed from the reference sentence. The remaining words, which represent the content, are used as input into the model, along with their POS tags and lemmas. 

In this way we encourage models to construct a sentence using content and style independently. This will allow us to vary the stylistic controls while keeping the content constant, and successfully perform style transfer. When generating a new sentence, the controls correspond to the counts of the corresponding syntactic features that we expect to be realized in the output.

\begin{figure*}
\centering
\includegraphics[scale=.55]{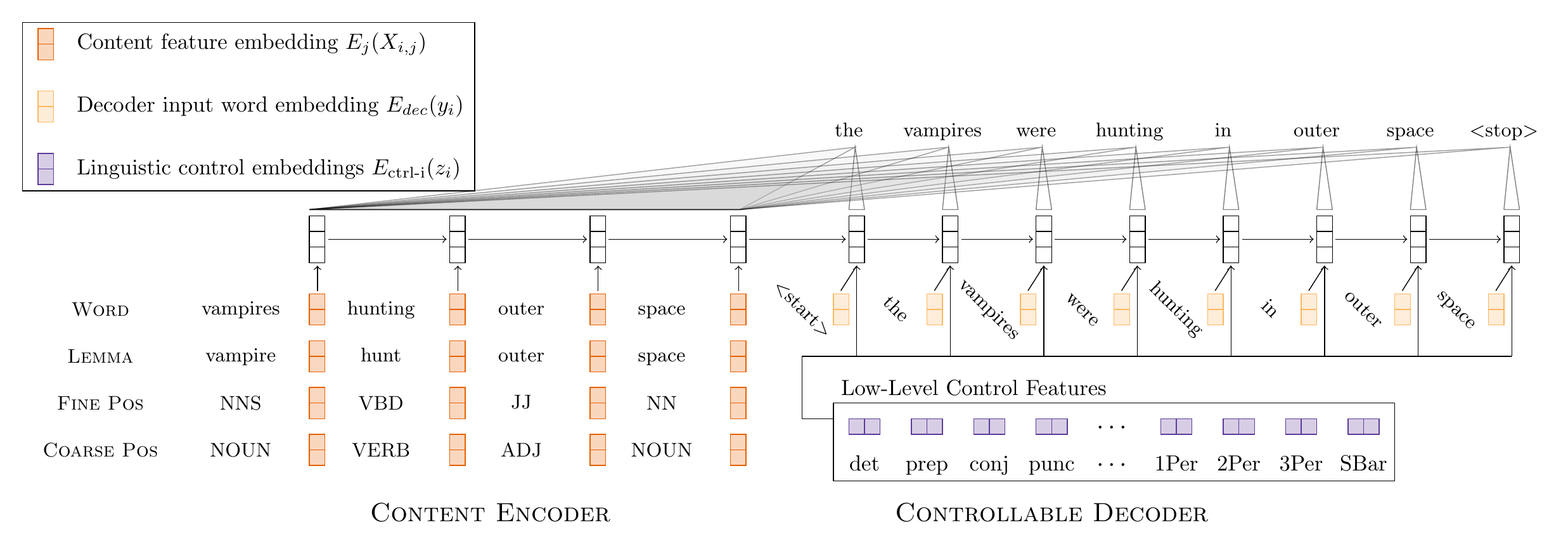}
\caption{A schematic depiction of our style control model.}
\label{fig:model}
\end{figure*}

\subsection{Neural Architecture}

We implement our feature controlled language model
using a neural encoder-decoder with attention \cite{bahdanau2014neural}, using  
2-layer uni-directional gated recurrent units (GRUs)
for the encoder and decoder \cite{cho2014learning}.

The input to the encoder is a sequence of $M$
content words, along with their lemmas, 
and fine and coarse grained part-of-speech (POS) tags,\footnote{We use the Penn Treebank \cite{marcus1994penn} 
and Universal Dependencies \cite{deuniversal} tagsets
for the fine and coarse-grained POS respectively.}
i.e. $X_{.,j} = (x_{1,j},\ldots,x_{M,j})$ for $j 
\in \mathcal{T} = \{\textrm{word, lemma, fine-pos, coarse-pos}\}$.
We embed each token (and its lemma and POS) before
concatenating, and feeding into the encoder GRU
to obtain encoder hidden states, $
c_i = \operatorname{gru}(c_{i-1}, \left[E_j(X_{i,j}),
 \; j\in \mathcal{T} \right]; \omega_{enc}) 
 $ for $i \in {1,\ldots,M},$ where 
 initial state $c_0$, encoder GRU
 parameters $\omega_{enc}$ and embedding matrices
 $E_j$ are learned parameters.
 
 The decoder sequentially generates 
 the outputs, i.e. a sequence of $N$ tokens $y =(y_1,\ldots,y_N)$, where all tokens $y_i$ are drawn
 from a finite output vocabulary $\mathcal{V}$.
 To generate the each token we first
 embed the previously generated token $y_{i-1}$ and a vector of $K$ control features $z = (
 z_1,\ldots, z_K)$ (using embedding matrices $E_{dec}$ and $E_{\textrm{ctrl-1}}, \ldots, E_{\textrm{ctrl-K}}$ respectively),
 before concatenating them into a vector $\rho_i,$ and feeding them into the decoder side GRU
 along with the previous decoder state $h_{i-1}$:
 \begin{align*}
  \rho_i& = \left[E_{dec}(y_{i-1}), E_{\textrm{ctrl-1}}(z_1), \ldots, E_{\textrm{ctrl-K}}(z_K) \right]\\
   h_i &= \operatorname{gru}\left(h_{i-1}, \rho_i;\;\omega_{dec}\right),
 \end{align*}
 where $\omega_{dec}$ are the decoder side GRU parameters.
 
 Using the decoder hidden state $h_i$ we then attend to the encoder
 context vectors $c_j$, computing attention scores $\alpha_{i,j}$, where \begin{align*}
 a_{i,j} =& \nu^{\intercal}\tanh\left(W^{\intercal} \left[ \begin{array}{l} c_j \\ h_i \end{array} \right]\right)\\ 
 \alpha_{i,j} =& \frac{\exp\left\{ a_{i,j} \right\}}{\sum_{j^\prime} \exp\{a_{i,j^\prime}\}},\end{align*}
 before passing $h_i$ and the attention weighted 
 context $\bar{c}_i=\sum_{j=1}^M \alpha_{i,j} c_j$ into a single hidden-layer perceptron with 
 softmax output to compute the next token prediction probability,
 \begin{align*}
      o_i  = &\tanh\left(U^\intercal \left[
        \begin{array}{l}h_i\\ \bar{c}_i \end{array}\right] + u \right)\\
    p(y_i|y_{<i},X)  \propto & \exp\left\{V_{y_i}^\intercal o_i + v_{y_i}\right\}.
    \end{align*}
 where $W,U,V$ and $u,v, \nu$ are parameter matrices and vectors 
 respectively. 
 
 Crucially, the controls $z$ remain fixed for all input decoder steps.
 Each $z_k$ represents the frequency of one of the low-level features described in 
 \autoref{sec:formalstyle}. During training on the reconstruction task, we can observe the full output sequence $y,$ and so 
 we can obtain counts for each control feature directly. Controls receive a different embedding depending on their frequency, where counts of 0-20 each get a unique embedding, and counts 
 greater than 20 are assigned to the same embedding. At test time, we set the values of the
 controls according to procedure described in Section~\ref{setfeats}.

 We use embedding sizes of 128, 128, 64, and 32 for  token, lemma, fine, and coarse grained POS embedding matrices respectively.  Output token embeddings $E_{dec}$ have size 512, and 50 for the control feature
 embeddings.
 We set 512 for all GRU and perceptron output sizes. We refer to this model as the
 StyleEQ model.\footnote{We think of the suite of feature controls as knobs akin to a parametric equalizer (EQ) on a HiFi-stereo.} See \autoref{fig:model}
 for a visual depiction of the model.\footnote{Implementation code can be found at:\\ \url{https://github.com/kedz/styleeq}}

  \subsubsection{Baseline Genre Model} We compare the above model to a similar model, where
 rather than explicitly represent $K$ features as input, we have $K$ features
 in the form of a genre embedding, i.e. we learn a genre specific embedding
 for each of the gothic, scifi, and philosophy genres, as studied in \citet{fu2018style} and \citet{zhaoetal}. To generate in a specific
 style, we simply set the appropriate embedding. We use genre embeddings of
 size 850 which is equivalent to the total size of the $K$ feature embeddings
 in the StyleEQ model.
 
 \subsubsection{Training} We train both models with minibatch stochastic gradient descent
 with a learning rate of 0.25, weight decay penalty of 0.0001, and batch 
 size of 64. We also apply 
dropout with a drop rate of 0.25 to all embedding layers, the GRUs, and preceptron
hidden layer. We train for a maximum of 200 epochs, using validation set 
BLEU score \cite{papineni2002bleu} to select the final model iteration for evaluation.
 
\subsubsection{Selecting Controls for Style Transfer}
\label{setfeats}

In the Baseline model, style transfer is straightforward: given an input sentence in one style, fix the encoder content features while selecting a different genre embedding. In contrast, the StyleEQ model requires selecting the counts for each control. Although there are a variety of ways to do this, we use a method that encourages a diversity of outputs.

In order to ensure the controls match the reference sentence in magnitude, we first find all sentences in the target style with the same number of words as the reference sentence. Then, we add the following constraints: the same number of proper nouns, the same number of nouns, the same number of verbs, and the same number of adjectives. 
We randomly sample $n$ of the remaining sentences, and for each of these `sibling' sentences, we compute the controls. For each of the new controls, we generate a sentence using the original 
input sentence content features.
The generated sentences are then reranked using the length normalized log-likelihood under the model. We can then select the highest scoring sentence as our style-transferred output, or take the top-$k$ when we need a diverse set of outputs.

The reason for this process is that although there are group-level distinctive controls for each style, e.g. the high use of punctuation in philosophy books or of first person pronouns in gothic novels, at the sentence level it can understandably be quite varied. This method matches sentences between styles, capturing the natural distribution of the corpora.

\section{Automatic Evaluations}

\subsection{BLEU Scores \& Perplexity}
In \autoref{tab:blueperpl} we report BLEU scores for the reconstruction 
of test set sentences from their content and feature representations,
as well as the model perplexities of the reconstruction. For both models, we use beam decoding
with a beam size of eight. Beam candidates are ranked according to their length
normalized log-likelihood.
On these automatic measures we see that StyleEQ is better able to reconstruct the original 
sentences. In some sense this evaluation is mostly a sanity check, as the 
feature controls contain more locally specific information than the genre
embeddings, which say very little about how many specific function words 
one should expect to see in the output.

\begin{table}[t]
\centering
\begin{tabular}{lrr}
\toprule
Model & BLEU & Perplexity\\
\midrule
Baseline & 25.07 & 4.60\\
StyleEQ & \textbf{30.04} & \textbf{3.33}\\
\bottomrule
\end{tabular}
\caption{Test set reconstruction BLEU score and perplexity (in nats).}
\label{tab:blueperpl}
\end{table}

\subsection{Feature Control}
\label{section:feat_ctrl}

\begin{table}
\center
\begin{tabular}{lrrr}
\toprule
Control &  Exact &  Direction &  Atomic \\
\midrule
S             &  18.99 &      43.34 &   23.86 \\
SBAR          &  24.22 &      41.41 &   18.16 \\
ADVP          &  20.78 &      27.65 &   21.96 \\
FRAG          &  24.47 &      26.60 &   19.71 \\
conjunction   &  93.56 &      98.75 &   11.43 \\
determiner    &  81.11 &      95.67 &   16.98 \\
3rdNeutralPer &  40.70 &      78.56 &    8.97 \\
3rdFemalePer  &  32.77 &      65.53 &   12.62 \\
3rdMalePer    &  36.20 &      75.72 &    9.27 \\
1stPer        &  79.47 &      94.48 &   12.80 \\
2ndPer        &  78.01 &      96.69 &   13.48 \\
3rdPer        &  29.08 &      70.92 &   10.56 \\
helperVerbs   &  69.92 &      90.23 &   12.30 \\
negation      &  68.85 &      93.21 &   12.88 \\
simple prep   &  49.32 &      77.74 &   19.86 \\
position prep &  47.18 &      79.42 &   19.42 \\
punctuation   &  84.83 &      91.71 &   13.05 \\
\bottomrule
\end{tabular}
\caption{Percentage rates of Exact,
Direction, and Atomic feature control changes. See \autoref{section:feat_ctrl} for
explanation.}
\label{table:autoeval:ctrl}
\end{table}

Designing controllable language models is often difficult because of the various dependencies between tokens;
when changing one control value
it may effect other aspects of the surface realization.
For example, increasing the number of conjunctions
may effect how the generator places prepositions to compensate for
structural changes in the sentence. Since our features are deterministically
recoverable, we can perturb an individual control value 
and check to see that the desired
change was realized in the output. Moreover, we can check the amount of change
in the other non-perturbed features to measure the independence of
the controls.

We sample 50 sentences from each genre from the test set.
For each sample, we create a perturbed control setting for each 
control by adding $\delta$ to the original control value.
This is done for $\delta \in \{-3, -2, -1, 0, 1, 2, 3\}$, skipping any
settings where the new control value would be negative.

\autoref{table:autoeval:ctrl} shows the results of this experiment.
The \emph{Exact} column displays the percentage of generated texts 
that realize the exact number of control features
specified by the perturbed control.
High percentages in the \emph{Exact} column indicate greater one-to-one correspondence between the control and surface realization. 
For example, if the input was ``Dracula and Frankenstein and the mummy,'' and we change the conjunction feature by $\delta=-1$, an output of ``Dracula, Frankenstein and the mummy,''
would count towards the \emph{Exact} category, while ``Dracula, Frankenstein, the mummy,'' would not.

The \emph{Direction} column specifies
the percentage of cases where the generated text produces a changed
number of the control features that, while not exactly matching 
the specified value of the perturbed control, does change from the original
in the correct direction. For example, if the input again was ``Dracula and Frankenstein and the mummy,'' and we change the conjunction feature by $\delta=-1$, both outputs of ``Dracula, Frankenstein and the mummy,''
and  ``Dracula, Frankenstein, the mummy,'' would count towards \emph{Direction}.
High percentages in \emph{Direction} mean that 
we could roughly ensure desired surface realizations by modifying
the control by a larger $\delta$.

Finally, the \emph{Atomic} column specifies the percentage of cases
where the generated text with the perturbed control only realizes changes
to that specific control, while other features remain constant. 
For example, if the input was ``Dracula and Frankenstein in the castle,'' and we set the conjunction feature to $\delta=-1$, an output of ``Dracula near Frankenstein in the castle,''
would not count as \emph{Atomic} because, while the number of conjunctions did decrease by one,
the number of simple preposition changed. An output of ``Dracula, Frankenstein in the castle,''
would count as \emph{Atomic}.
High percentages
in the \emph{Atomic} column indicate this feature is only loosely coupled
to the other features and can be changed without modifying other aspects
of the sentence.

Controls such as \textit{conjunction}, \textit{determiner}, and \textit{punctuation} are highly controllable, with \emph{Exact} rates above 80\%. But with the exception of the constituency parse features, all controls have high \emph{Direction} rates, many in the 90s. These results indicate our model successfully controls these features. The fact that the \emph{Atomic} rates are relatively low is to be expected, as controls are highly coupled -- e.g. to increase \emph{1stPer}, it is likely another pronoun control will have to decrease.

\begin{table*}[t]
\center
\begin{tabular}{ll|r|rrr|rrr|rrr}
\toprule
 \multicolumn{3}{r}{} & \multicolumn{3}{c}{scifi (s)} & \multicolumn{3}{c}{philosophy (p)} & \multicolumn{3}{c}{gothic (g)} \\
Model	&Method	&all	&s$\rightarrow$s	&s$\rightarrow$p	&s$\rightarrow$g	&p$\rightarrow$s	&p$\rightarrow$p	&p$\rightarrow$g	&g$\rightarrow$s	&g$\rightarrow$p	&g$\rightarrow$g \\
\midrule
Baseline	&all	&.424	&.639	&.344	&.301	&.242	&.818	&.140	&.483	&.422	&.437\\
Baseline	&top	&.429	&.666	&.344	&.301	&.242	&.819	&.140	&.483	&.422	&.400\\
Baseline	&oracle	&\textbf{.493}	&.851	&.344	&.301	&.242	&.940	&.140	&.483	&.422	&.750\\
\midrule
StyleEQ	    &all	&.413	&.561	&.348	&.322	&.167	&.803	&.268	&.378	&.467	&.426\\
StyleEQ	    &top	&.382	&.573	&.307	&.221	&.201	&.800	&.165	&.458	&.430	&.436\\
StyleEQ	    &oracle	&\textbf{.841}	&.804	&.834	&.947	&.560	&.926	&.900	&.866	&.914	&.679\\
\bottomrule
\end{tabular}
\caption{\textit{Ablated NVA} classifier accuracy using three different methods of selecting an output sentence. This is additionally split into the nine transfer possibilities, given the three source styles. The StyleEQ model produces far more diverse outputs, allowing the oracle method to have a very high accuracy compared to the Baseline model.}
\label{table:fasttext-results}
\end{table*}

\subsection{Automatic Classification}

For each model we look at the classifier prediction accuracy of reconstructed and transferred sentences. In particular we use the \textit{Ablated NVA} classifier, as this is the most content-blind one.

We produce 16 outputs from both the Baseline and StyleEq models. For the Baseline, we use a beam search of size 16. For the StyleEQ model, we use the method described in Section~\ref{setfeats} to select 16 `sibling' sentences in the target style, and generated a transferred sentence for each.\footnote{For each `sibling' we used a beam search of size 8 and selected the top candidate according to length normalized log-likelihood.}
We look at three different methods for selection: \textit{all}, which uses all output sentences; \textit{top}, which selects the top ranked sentence based on the score from the model; and \textit{oracle}, which selects the sentence with the highest classifier likelihood for the intended style. 

The reason for the third method, which indeed acts as an oracle, is that using the score from the model didn't always surface a transferred sentence that best reflected the desired style. Partially this was because the model score was mostly a function of how well a transferred sentence reflected the distribution of the training data. But additionally, some control settings are more indicative of a target style than others. The use of the classifier allows us to identify the most suitable control setting for a target style that was roughly compatible with the number of content words.  

In \autoref{table:fasttext-results} we see the results. Note that for both models, the \textit{all} and \textit{top} classification accuracy tends to be quite similar, though for the Baseline they are often almost exactly the same when the Baseline has little to no diversity in the outputs. 

However, the \textit{oracle} introduces a huge jump in accuracy for the StyleEQ model, especially compared to the Baseline, 
partially because the diversity of outputs from StyleEQ is much higher; often the Baseline model produces no diversity -- the 16 output sentences may be nearly identical, save a single word or two.
It's important to note that neither model uses the classifier in any way except to select the sentence from 16 candidate outputs. 

What this implies is that lurking within the StyleEQ model outputs are great sentences, even if they are hard to find. In many cases, the StyleEQ model has a classification accuracy above the base rate from the test data, which is 75\% (see \autoref{table:classifiers}).

\section{Human Evaluation}

\begin{table*}
\begin{center}
\small
\begin{tabular}{ l | p{7cm} | p{7cm} }
  \toprule
  Setting & StyleEQ output & Baseline output\\
  \midrule
  reference & Her face had turned beet red. & Her face had turned beet red. \\
  s$\rightarrow$s & her face had turned thereto red. & his face had turned out of the dissolution of the red.\\
  s$\rightarrow$g & her face had turned to me, the realization red. & her face had turned, and, with a modesty of red.\\
  s$\rightarrow$p & in the face, had turned--that was, the realization red. & his face had turned, and, with a modesty of red. \\
  \midrule
  reference & The desire for exclusive markets is one of the most potent causes of war. & The desire for exclusive markets is one of the most potent causes of war.\\
  p$\rightarrow$p & the desire of exclusive markets is one of the most potent causes of war.  & the desire of exclusive markets is one of the most potent causes of war.\\
  p$\rightarrow$s & but his desire is an exclusive markets, one of the most potent causes of war. & the desire of the exclusive markets were one of the most potent causes of war.\\
  p$\rightarrow$g & i am a desire of your exclusive markets, and that you are one of the most potent causes of your war in me. & the desire of the exclusive markets were one of the most potent causes of war.\\
  \midrule
  reference & a little while, and all this will appear a dream. & a little while, and all this will appear a dream. \\
  g$\rightarrow$g & but a little while, all this will appear a dream.  & a little while all it would appear in a dream.\\
  g$\rightarrow$s & he wasn't a little while all he could appear in the dream. & a little while all it would appear in a dream. \\
  g$\rightarrow$p & a little while--all that would appear to do, dream. & a little while all will appear in a dream. \\
  \bottomrule
\end{tabular}
\normalsize
\caption[]{\label{table:cherrypicking} Example outputs (manually selected) from both models. The StyleEQ model successfully rewrites the sentence with very different syntactic constructions that reflect style, while the Baseline model rarely moves far from the reference.}
\end{center}
\end{table*}

\autoref{table:cherrypicking} shows example outputs for the StyleEQ and Baseline models\footnote{The outputs are manually selected from the set of 16 candidate output sentences.}. Through inspection we see that the StyleEQ model successfully changes syntactic constructions in stylistically distinctive ways, such as increasing syntactic complexity when transferring to philosophy, or changing relevant pronouns when transferring to sci-fi. In contrast, the Baseline model doesn't create outputs that move far from the reference sentence, making only minor modifications such changing the type of a single pronoun.

To determine how readers would classify our transferred sentences, we recruited three English Literature PhD candidates, all of whom had passed qualifying exams that included determining both genre and era of various literary texts.

\subsection{Fluency Evaluation}

To evaluate the fluency of our outputs, we had the annotators score reference sentences, reconstructed sentences, and transferred sentences on a 0-5 scale, where 0 was incoherent and 5 was a well-written human sentence. 

\autoref{table:fluency} shows the average fluency of various conditions from all three annotators. Both models have fluency scores around 3. Upon inspection of the outputs, it is clear that many have fluency errors, resulting in ungrammatical sentences. 

Notably the Baseline often has slightly higher fluency scores than the StyleEQ model. This is likely because the Baseline model is far less constrained in how to construct the output sentence, and upon inspection often reconstructs the reference sentence even when performing style transfer. In contrast, the StyleEQ is encouraged to follow the controls, but can struggle to incorporate these controls into a fluent sentence.

The fluency of all outputs is lower than desired. 
We expect that incorporating pre-trained language models would increase the fluency of all outputs without requiring larger datasets.

\begin{table}
\center
\begin{tabular}{llccc}
\toprule
& & \multicolumn{3}{c}{fluency}\\
Sentence Type & Model       & A1  & A2  & A3 \\
\midrule
Reference & none    & 4.94 & 4.47 & 4.82 \\
\noalign{\smallskip}
Reconstruction & Baseline    & 3.48 & 3.09 & 4.13 \\
        & StyleEQ           & 3.60 & 2.93 & 3.96 \\
\noalign{\smallskip}
Transferred & Baseline      & 3.36 & 4.17 & 3.30 \\
        &StyleEQ        & 3.22 & 3.86 & 3.00 \\
\bottomrule
\end{tabular}
\caption{\label{table:fluency} Fluency scores (0-5, where 0 is incoherent) of sentences from three annotators. The Baseline model tends to produce slightly more fluent sentences than the StyleEQ model, likely because it is less constrained.}
\end{table}

\subsection{Human Classification}

Each annotator annotated 90 reference sentences 
(i.e. from the training corpus) 
with which style they thought the sentence was from. The accuracy on this baseline task for annotators A1, A2, and A3 was 80\%, 88\%, and 80\% respectively, giving us an upper expected bound on the human evaluation. 

In discussing this task with the annotators, they noted that content is a heavy predictor of genre, and that would certainly confound their annotations. To attempt to mitigate this, we gave them two annotation tasks: \textit{which-of-3} where they simply marked which style they thought a sentence was from, and \textit{which-of-2} where they were given the original style and marked which style they thought the sentence was transferred into.

For each task, each annotator marked 180 sentences: 90 from each model, with an even split across the three genres. Annotators were presented the sentences in a random order, without information about the models. In total, each marked 270 sentences. (Note there were no reconstructions in this annotation task.)

\autoref{table:humanclassifiers} shows the results. In both tasks, accuracy of annotators classifying the sentence as its intended style was low. In \textit{which-of-3}, scores were around 20\%, below the chance rate of 33\%. In \textit{which-of-2}, scores were in the 50s, slightly above the chance rate of 50\%. This was the case for both models.
There was a slight increase in accuracy for the StyleEQ model over the Baseline for \textit{which-of-3}, but the opposite trend for \textit{which-of-2}, suggesting these differences are not significant.

It's clear that it's hard to fool the annotators. Introspecting on their approach, the annotators expressed having immediate responses based on key words -- for instance any references of `space' implied `sci-fi'. We call this the `vampires in space' problem, because no matter how well a gothic sentence is rewritten as a sci-fi one, it's impossible to ignore the fact that there is a vampire in space.
The transferred sentences, in the eyes of the \textit{Ablated NVA} classifier (with no access to content words), did quite well transferring into their intended style. But people are not blind to content.

\begin{table}
\center
\begin{tabular}{lrrrrrr}
\toprule
& \multicolumn{3}{c}{\textit{which-of-3}} & \multicolumn{3}{c}{\textit{which-of-2}}\\
Model       & A1  & A2  & A3  & A1  & A2  & A3\\
\midrule
Baseline    & .21 & .17 & .17 & .57 & .51 & .58\\
StyleEQ     & .24 & .20 & .17 & .54 & .51 & .48\\
\bottomrule
\end{tabular}
\caption{\label{table:humanclassifiers} Accuracy of three annotators in selecting the correct style for transferred sentences. In this evaluation there is little difference between the models.}
\end{table}

\subsection{The `Vampires in Space' Problem}

Working with the annotators, we regularly came up against the 'vampires in space' problem: while syntactic constructions account for much of the distinction of literary styles, these constructions often co-occur with distinctive content.

Stylometrics finds syntactic constructions are great at fingerprinting, but suggests that these constructions are surface realizations of higher-level stylistic decisions. The number and type of personal pronouns is a reflection of how characters feature in a text. A large number of positional prepositions may be the result of a writer focusing on physical descriptions of scenes. 
In our attempt to decouple these, we create Frankenstein sentences, which piece together features of different styles -- we are putting vampires in space. 

Another way to validate our approach would be to select data that is stylistically distinctive but with similar content: perhaps genres in which content is static but language use changes over time, stylistically distinct authors within a single genre, or parodies of a distinctive genre.

\section{Conclusion and Future Work}

We present a formal, extendable model of style that can add control to any neural text generation system. 
We model style as a suite of low-level linguistic controls, 
and train a neural encoder-decoder model to reconstruct reference sentences given only content words and the setting of the controls. 
In automatic evaluations, we show that our model can fool a style classifier 84\% of the time and outperforms a baseline genre-embedding model. In human evaluations, we encounter the `vampires in space' problem in which content and style are equally discriminative but people focus more on the content. 

In future work we would like to model higher-level syntactic controls. \citet{scaleofsentence} show that differences in clausal constructions, for instance having a dependent clause before an independent clause or vice versa, is a marker of style appreciated by the reader. Such features would likely interact with our lower-level controls in an interesting way, and provide further insight into style transfer in text.

\section*{Acknowledgements}

Katy Gero is supported by an NSF GRF (DGE - 1644869).
We would also like to thank Elsbeth Turcan for her helpful comments.

\bibliography{citations}
\bibliographystyle{acl_natbib}

\end{document}